\documentclass{ecai}
\usepackage{graphicx}
\usepackage{latexsym}
\usepackage{amsmath}
\usepackage{amssymb}
\usepackage{booktabs}
\usepackage{times}
\usepackage{soul}
\usepackage{url}
\usepackage[colorlinks,linkcolor=black,anchorcolor=black,citecolor=black]{hyperref}
\usepackage[utf8]{inputenc}
\usepackage{amsmath}
\usepackage{amsthm}
\usepackage{amsfonts}
\usepackage{booktabs}
\usepackage{algorithm}
\usepackage{algorithmic}
\usepackage[switch]{lineno}
\usepackage{color}
\usepackage{multirow}
\usepackage{float}
\usepackage{colortbl}
\usepackage[table,xcdraw]{xcolor}

\begin{document}

\begin{frontmatter}

\title{PMAA: A Progressive Multi-scale Attention Autoencoder Model for High-performance Cloud Removal from Multi-temporal Satellite Imagery}

\author[A]{\fnms{Xuechao}~\snm{Zou}}
\author[B]{\fnms{Kai}~\snm{Li}\thanks{Xuechao Zou and Kai Li contributed equally to this work.}}
\author[B]{\fnms{Junliang}~\snm{Xing}}
\author[A,B]{\fnms{Pin}~\snm{Tao}\thanks{Corresponding Author. Email: taopin@tsinghua.edu.cn.}}
\author[A]{\fnms{Yachao}~\snm{Cui}} 

\address[A]{Department of Computer Technology and Applications, Qinghai University, Xining, China}
\address[B]{Department of Computer Science and Technology, Tsinghua University, Beijing, China}


\begin{abstract}
Satellite imagery analysis plays a pivotal role in remote sensing; however, information loss due to cloud cover significantly impedes its application. Although existing deep cloud removal models have achieved notable outcomes, they scarcely consider contextual information. This study introduces a high-performance cloud removal architecture, termed Progressive Multi-scale Attention Autoencoder (PMAA), which concurrently harnesses global and local information to construct robust contextual dependencies using a novel Multi-scale Attention Module (MAM) and a novel Local Interaction Module (LIM). PMAA establishes long-range dependencies of multi-scale features using MAM and modulates the reconstruction of fine-grained details utilizing LIM, enabling simultaneous representation of fine- and coarse-grained features at the same level. With the help of diverse and multi-scale features, PMAA consistently outperforms the previous state-of-the-art model CTGAN on two benchmark datasets. Moreover, PMAA boasts considerable efficiency advantages, with only 0.5\% and 14.6\% of the parameters and computational complexity of CTGAN, respectively. These comprehensive results underscore PMAA's potential as a lightweight cloud removal network suitable for deployment on edge devices to accomplish large-scale cloud removal tasks. \textit{Our source code and pre-trained models are available at \url{https://github.com/XavierJiezou/PMAA}}.
\end{abstract}

\end{frontmatter}

\section{Introduction}
\label{intro}


With the rapid development of remote sensing technologies, satellite imagery has been widely applied in various fields, such as grassland monitoring~\cite{grasslandmonitoring}, ground target detection~\cite{objectdetection2}, land cover classification~\cite{landcover1,landcover2}, \textit{etc}. Nonetheless, due to weather conditions, clouds often disrupt the imaging process of optical sensors carried on satellites, resulting in information loss and image quality degradation. Cloud removal from satellite imagery attempts to reconstruct the original information in cloud-covered areas to solve the above problems. It is a critical preprocessing step that significantly affects the effective use of satellite imagery.

Recently, convolutional neural networks (CNNs~\cite{deeplearning}) and generative adversarial networks (GANs~\cite{gans}) have shown significant improvements in cloud removal performance for satellite imagery. Among these, mono-temporal cloud removal methods~\cite{mcgan,8803666} generate a corresponding cloud-free image using a single cloudy image. These methods exhibit stable performance in cloud removal from satellite imagery. However, when cloud coverage is extensive, sufficient information cannot be effectively obtained, making it challenging to generate cloud-free satellite imagery and often impossible.

\begin{figure}[t]
\centering
\includegraphics[width=\linewidth]{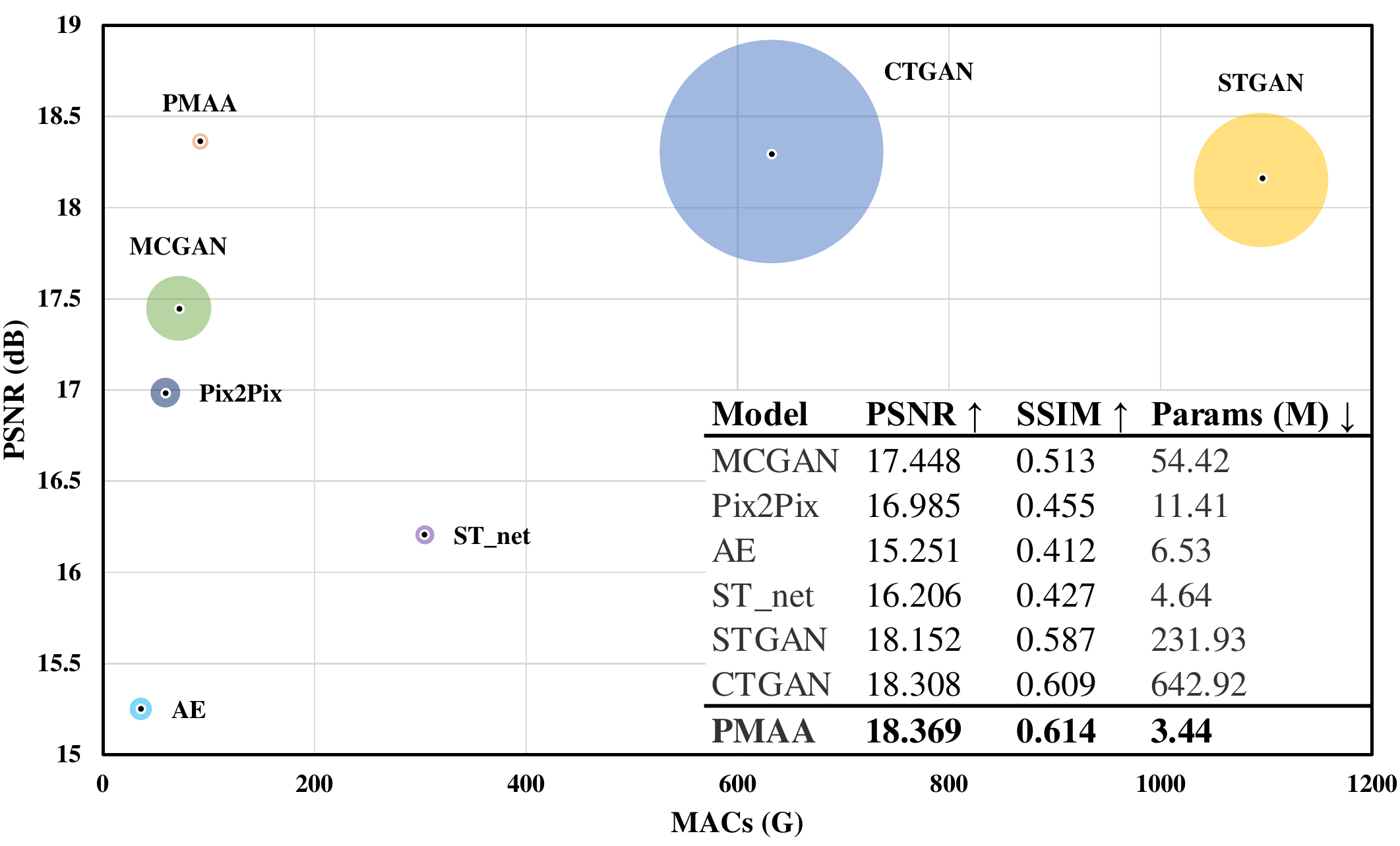}
\caption{Performance and efficiency comparison on the \emph{Sen2\_MTC\_New} dataset. Compared to existing methods, our proposed PMAA achieves the SOTA performance while being computationally efficient and having few parameters. The area of the circle indicates the number of model parameters. Complete results are shown in Table~\ref{tab:pcr}.}
\label{sota}
\end{figure}

As remote sensing technology advances, satellite revisitation periods to the same location become increasingly shorter. We can quickly obtain multiple images of the same area captured by satellites at different times. Recently, some researchers have explored utilizing multi-temporal satellite images and successfully applied them to cloud removal to enhance performance. The multi-temporal cloud removal method uses multiple cloudy images at the same location as inputs and generates a cloud-free image using spatial and temporal information. Among these, \cite{ae} proposed a CNN-based autoencoder using multi-temporal satellite imagery to remove clouds. \cite{stnet} combined cloud detection network to improve performance. \cite{stgan} treated the cloud removal problem as a challenge of conditional image synthesis and proposed a spatiotemporal generative network for cloud removal. \cite{ctgan} proposed a Transformer-based GAN for cloud removal, increasing the accuracy of generated cloud-free images.

However, cloud removal methods suffer from more issues. Firstly, almost all methods do not consider the model complexity, potentially resulting in additional time consumption, especially for large-resolution satellite imagery. Secondly, existing methods~\cite{mcgan,pix2pix,ae,stnet,stgan,ctgan} do not consider progressively generating cloud-free images but only perform a single stage of reconstruction, which may affect the reconstruction of spatial details in the image~\cite{mprnet}. Thirdly, they generally only consider the impact of local information on generating cloud-free images without considering the combination of global and local information. Therefore, the challenge remains to efficiently exploit global and local information for the cloud removal task.

To address the mentioned issues, we present a high-performance Progressive Multi-scale Attention Autoencoder (PMAA) that effectively captures fine- and coarse-grained features across different scales, as depicted in Figure~\ref{pmaa}. The PMAA mainly comprises two novel components, the Multi-scale Attention Module (MAM) and the Local Interaction Module (LIM). Firstly, the MAM aggregates multi-scale features to acquire global attention through multi-temporal satellite images and then modulate the multi-scale features. Secondly, we use the LIM to connect the local and global features extracted by the MAM, which reconstructs the fine-grained image structure. To further enhance the fidelity of the generated images, we propose a novel progressive learning approach for iterative refinement. The main contributions of our work can be summarized in three key aspects.
\begin{itemize}
    \item We propose a novel lightweight architecture, named PMAA, for removing clouds from satellite imagery, which mainly incorporates two carefully designed components: MAM and LIM. 
    \item We design an efficient MAM that effectively compensates for the loss of accuracy in reconstructing cloud-free images by aggregating features of different spatial resolutions into global features.
    \item We design a new reconstruction component, LIM, to enhance the fine-grained generation of cloud-free images through selectively fusing global and local features.
\end{itemize}

Based on the above main contributions, we obtain a high-performance model for cloud removal from satellite imagery. Our extensive experiments on the PMAA demonstrate its superiority in the task of cloud removal. Specifically, PMAA consistently achieves state-of-the-art (SOTA) performance on the \emph{Sen2\_MTC\_Old}~\cite{stgan} and \emph{Sen2\_MTC\_New}~\cite{ctgan} datasets compared to previous methods. Furthermore, we also demonstrate the efficiency of PMAA, which achieves the optimal performance-efficiency trade-off. Compared to the previous SOTA model, PMAA saves approximately 99.5\% of parameters and 85.4\% of the computational cost, as shown in Figure~\ref{sota}.

\section{Related work}
\subsection{Cloud removal}
With the development of deep learning, cloud removal methods for satellite imagery have increasingly become the focus of current research. Existing methods can be classified into two types: mono-temporal and multi-temporal. The mono-temporal-based methods~\cite{mcgan,8803666,dip} will have a much faster inference time because its input uses only a single cloudy image. \textit{Enomoto et al.}~\cite{mcgan} proposed a CGAN-based method to achieve thin cloud removal on multispectral remote sensing data. \textit{Lin et al.} published a mono-temporal dataset and used Pix2Pix~\cite{pix2pix} as the baseline for cloud removal. Pan introduced a spatial attention mechanism in GAN to enhance the information recovery of cloud regions to generate better quality. \textit{Czerkawski et al.}~\cite{dip} used an internal learning regime based on the deep image before obtaining the ability to inpaint the cloud-affected regions.

However, when many clouds cover the satellite image, the mono-temporal-based methods may not obtain precise results, which limits its application in practical scenarios. Multi-temporal-based methods~\cite{ae,stnet,stgan,ctgan} obtain better results than mono-temporal-based methods using multi-temporal cloudy images to reconstruct a single cloud-free image. \textit{Sintarasirikulchai et al.}~\cite{ae} proposed cloud removal using convolutional autoencoders by training on a multi-temporal remote sensing dataset. \textit{Chen et al.}~\cite{stnet} combined cloud detection techniques to perform cloud removal by fusing spatiotemporal features of multi-temporal data. \textit{Sarukkai et al.}~\cite{stgan} treated the cloud removal problem as a conditional image synthesis challenge and proposed a spatiotemporal generative network for cloud removal. \textit{Huang et al.}~\cite{ctgan} proposed a Transformer-based GAN for de-clouding. Although they recover high-quality cloud-free images, the application's inference process takes more time. In contrast, our proposed PMAA focuses on multi-temporal satellite images' global and local features and efficiently reconstructs cloud-free images. Furthermore, PMAA dramatically reduces the computational cost of the model, making practical applications possible.

\subsection{Attention mechanism}
Attention mechanism originates from natural language processing (NLP), such as language modeling~\cite{devlin2018bert,liu2019roberta,lan2019albert}, machine translation~\cite{transformer}, and generative tasks~\cite{afrcnn,tdanet}. They estimate the relationship between current and global features via the attention mechanism. Recently, the computer vision field has also successfully used attention mechanisms to improve model performance. \cite{zits} proposed using a transformer-based model to learn normalized grayscale sketch tensor space for accomplishing painting tasks. This attention-based model can significantly learn global structures with long-range dependencies, which helps to address the limitations of Convolutional Neural Networks (CNN) in recovering the overall structure of images. The UQ-Transformer \cite{Liu_2022_CVPR} takes unquantized feature vectors from the encoder as an input and uses the quantized tokens of unmasked patches as prediction targets, thereby reducing information loss and improving prediction accuracy. However, these methods always transfer the attention map to deeper layers, which can lead to shallow features being less affected by attention and bring limited performance improvements. In addition, existing methods~\cite{vit,swin} tend to divide the image into non-overlapping patches to reduce the computational cost, which is unreliable for cloud removal. This is because the image's cloud distribution is not uniform, leading to inconsistent cloud occupancy in different patches.


\begin{figure}[tb]
\centering
\includegraphics[width=\linewidth]{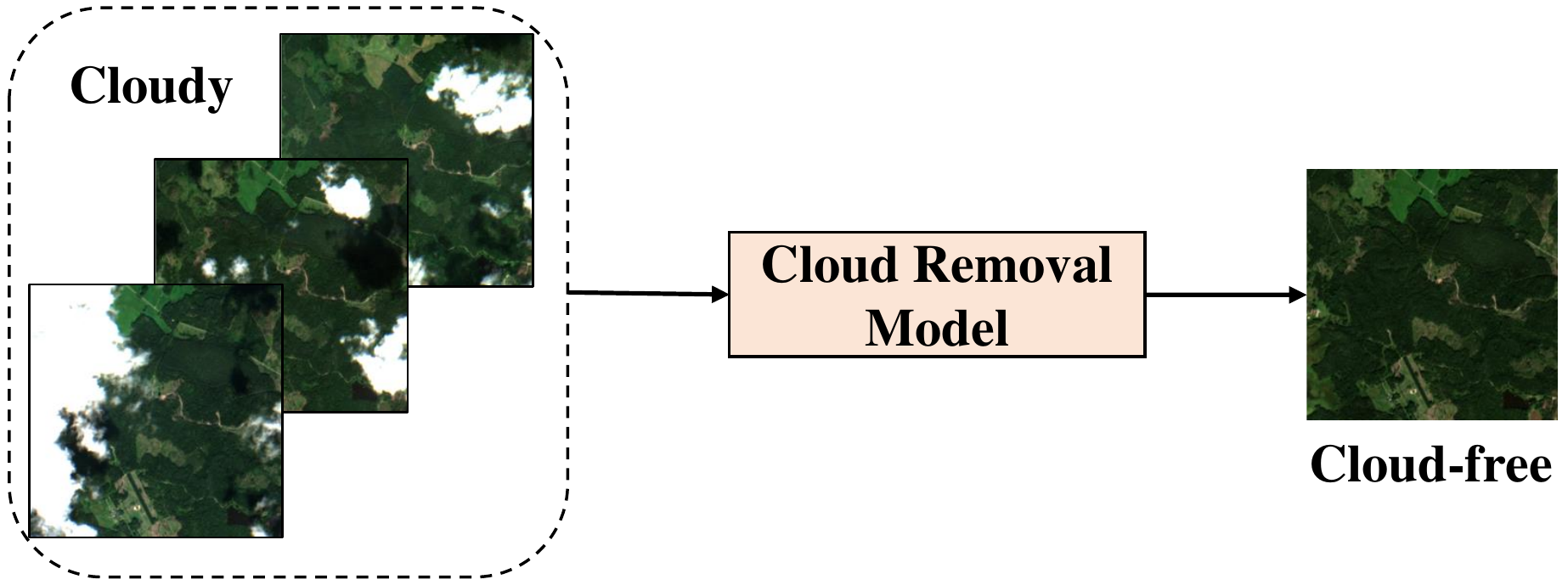}
\caption{A pipeline for cloud removal task using multi-temporal satellite imagery. Input is multiple cloudy images of the same location and adjacent moments, and output is a cloud-free image of the corresponding location.}
\label{fig:pipeline}
\end{figure}

\subsection{Progressive learning}
Progressive learning aims to split complex processes (\textit{e.g.}, direct reconstruction of cloud-free images) into multiple easier and smaller stages (\textit{e.g.}, multi-step reconstruction of cloud-free images) to improve model performance and is widely considered in visual tasks and speech tasks, such as image synthesis~\cite{ddpm}, image super-resolution~\cite{wang2018fully,survey-super-resolution}, and speech separation \cite{afrcnn,tdanet}. In addition, progressive learning is more in line with how humans perceive images because the human visual system does not process the whole scene simultaneously. Instead, it gradually focuses its attention on the part of the interest of the image and ignores the irrelevant details. It can combine information from different regions to reconstruct the complete scene in the brain \cite{miyawaki2008visual,naselaris2009bayesian,nishimoto2011reconstructing}.

\begin{figure*}[ht]
\centering
\includegraphics[width=\linewidth]{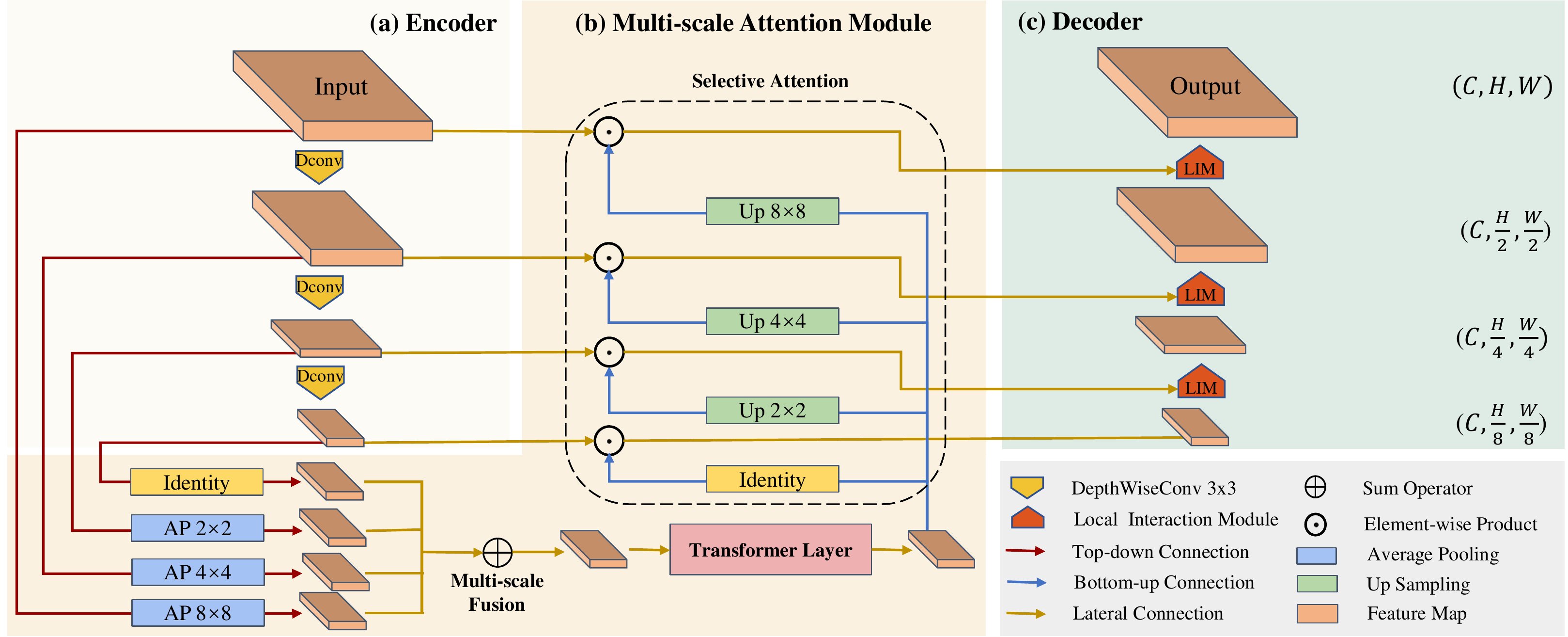}
\caption{Overview of our proposed high-performance cloud removal autoencoder. In the encoder, we downsample the input image $N$ times. Then, the multi-scale features are fused by averaging pooling and summation operations. A simplified transformer layer processes the fused features to obtain global attention, which is used to modulate the multi-scale features. In the reconstruction process (decoder), we use the local interaction module to recover more details.}
\label{pmaa}
\end{figure*}

\section{Method}

\subsection{Overall pipeline}

As shown in Figure~\ref{fig:pipeline}, the algorithm of multi-temporal cloud removal is expected to generate a cloud-free satellite image from multiple cloudy satellite images (same location, adjacent time). We denote three cloudy satellite images as $\{\mathbf{X}_{i} \in \mathbb{R}^{4 \times H \times W}|i=1,2,3\}$, where $H$ and $W$ are the image's height and width, respectively, and ``4" denotes the four Spectral channels (RGB and infra-red). And we denote a cloud-free image at the current location as $\mathbf{y} \in \mathbb{R}^{4 \times H \times W}$. We assume that for an arbitrary location, $\mathbf{X}_{i}$ changes slowly over time and that the cloud cover position in the image varies. 

Given the input multi-temporal satellite images $\{\mathbf{X}_{i}|i=1,2,3\}$, we firstly preprocess them using a weight shared bottleneck consisting of several convolutions with residual connection, which serve as input for the cloud removal autoencoder. Secondly, the encoder in the cloud removal autoencoder downsamples through convolutional layers with the stride size of $2\times2$, producing features with different spatial resolutions. Then they are aggregated to obtain fine- and coarse-grained representations. Thirdly, aggregated multi-scale features are fed into the MAM to obtain global attention and modulate multi-scale features. Then, the LIM in the decoder connects local and global features to reconstruct fine-grained image structures. Finally, we use a novel progressive learning method to cycle PMAA's cloud removal autoencoder to generate approximate cloud-free images.

\subsection{High-performance cloud removal autoencoder} \label{sec:crm}
\label{sec:autoencoder}

We design a novel high-performance cloud removal autoencoder (as shown in Figure~\ref{pmaa}) that receives the spatiotemporal features $\mathbf{U}_{c} \in \mathbb{R}^{12 \times H \times W}$ as input, which are obtained by concatenating $\{\mathbf{U}_i|i=1,2,3\}$ on the channel dimension. After progressive refinement, the cloud removal module ultimately yields a cloud-free image $\Bar{\mathbf{y}}$ at the current location. The cloud removal module is described in the following sections.

\subsubsection{Encoder}

To consider data dimensionality reduction and mimicking human cognitive processes, we commonly deepen image features through downsampling while preserving as much relevant information as possible. To this end, we introduce an encoder (Figure~\ref{pmaa}(a)) capable of extracting multi-scale features. To achieve this, we employ multiple depth-wise separable convolutional layers~\cite{mobilenets} with a kernel size of $3 \times 3$ and stride of $2 \times 2$ for reducing spatial image scale and increasing receptive field. After downsampling $N$ times, we obtain $N+1$ multi-scale features $\mathbf{F}_{i} \in \mathbb{R}^{C \times {\frac{H}{2^{i}}} \times {\frac{W}{2^{i}}} }|i=0,...,N\}$ with the incremental receptive field, where $C$ denotes the features channels. All convolutional layers are followed by instance normalization and ReLU activation function~\cite{relu}. Finally, the multi-scale features $\mathbf{F}_{i}$ are fed into the following Multi-scale Attention Module.

\subsubsection{Multi-scale attention module}
To endow the model with the capability of perceiving both fine- and coarse-grained visual features, we introduce a multi-scale attention module (MAM, Figure~\ref{pmaa}(b)) composed of three components: 1) Multi-scale Fusion; 2) Transformer Layer; 3) Selective Attention. The MAM works as follows.

\textbf{Multi-scale fusion}. First, we craft a multi-scale fusion approach without any parameters and with insignificant additional computation costs. It compresses all features $\{\mathbf{F}_{i}|i=0,...,N\}$ with the scale $(\frac{H}{2^{i}} \times {\frac{W}{2^{i}}})$ to the uniform scale $(\frac{H}{2^N},\frac{W}{2^N})$ using the adaptive average pooling layers and then fuse them by a summation operation to obtain a multi-scale representation 
\begin{equation}
    \mathbf{F}_\text{ms}=\sum_{i-0}^N H(\mathbf{F}_{i}),
\end{equation}
where $H(\cdot)$ represents adaptive average pooling layer. Finally, we use the $\mathbf{F}_\text{ms}$ as the input to the transformer layer in the next stage.

\textbf{Transformer layer}. Limited by the receptive fields of CNNs, some methods~\cite{pix2pix,mcgan,stgan} struggle to acquire enough contextual information for cloud removal, particularly when cloud cover is extensive. This is because an individual pixel and its receptive field may all fall in the cloudy region, preventing it from noticing information outside the cloud area and severely hindering the recovery of the cloud-free image. Therefore, we introduce self-attention~\cite{transformer} to encode spatial information to establish long-range dependency. So far, there are many vision transformers based on self-attention, such as ViT~\cite{vit} and Swin Transformer~\cite{swin}. To balance performance and efficiency, we adopt a simple self-attention implemented by using a convolutional modulation operation that uses large kernel convolution to avoid the problem of time-consuming and complex computation of the attention matrix (see Figure~\ref{fig:transformer+lim})
\begin{equation}
    \mathbf{F}_{a}=\mathbf{F}_\text{ms}+\alpha \mathbf{W}_{3}(\text{DConv}_{k \times k}(\mathbf{W}_{1}\mathbf{F}_\text{ms})\odot \mathbf{W}_{2}\mathbf{F}_\text{ms}),
\end{equation}
where $\mathbf{W}_{1},\mathbf{W}_{2},\mathbf{W}_{3}$ are linear layers and $\alpha$ is a learnable parameter, and $\text{DConv}_{k \times k}$ denotes depth-wise convolutional layer with a large kernel size. Then, a residual connection~\cite{resnet} is added after self-attention to reduce information loss. The self-attention is immediately followed by a feed-forward network (FFN), which consists of one depth-separable convolution layer and two linear layers.
\begin{equation}
    \mathbf{F}_{g}=\mathbf{F}_{a}+\beta \mathbf{V}_{2}(\mathbf{V}_{1}\mathbf{F}_{a}+\text{DConv}_{k \times k}(\mathbf{V}_{1}\mathbf{F}_{a})),
\end{equation}
where $\mathbf{V}_{1},\mathbf{V}_{2}$ are linear layers and $\beta$ is a learnable parameter. In summary, we obtain a feature $\mathbf{F}_{g}$ with global information by processing a transformer layer.

\textbf{Selective attention}. We use $\mathbf{F}_{g}$ as global attention to perform adaptive feature recalibration on $\mathbf{F}_{i}$ before integrating it into the decoder. This is because the earlier neural network layers possess rich low-level texture features, while the deeper layers have high-level semantic information. Specifically, we upsample $\mathbf{F}_{g}$ through nearest-neighbor interpolation to obtain the same spatial dimension as $\mathbf{F}_{i}$. Then, we obtain the modulated feature $\{\mathbf{F}_{i}' \in \mathbb{R}^{c \times {\frac{H}{2^{i}}}  \times {\frac{W}{2^{i}}} }|i=0,...,N\}$ through affine transformation
\begin{equation}
    \mathbf{F}_{i}'=\phi(\sigma(\mathbf{Z}_{1}(\mathbf{F}_{g}))) \odot \mathbf{Z}_{2}(\mathbf{F}_{i})+\phi(\mathbf{Z}_{3}(\mathbf{F}_{g})),
\end{equation}
where $\mathbf{Z}_{1},\mathbf{Z}_{2},\mathbf{Z}_{3}$ are linear layers, $\sigma$ represents the sigmoid activation function, and $\phi$ represents the nearest-neighbor interpolation.

\subsubsection{Decoder}

\begin{figure*}[t]
\centering
\includegraphics[width=\linewidth]{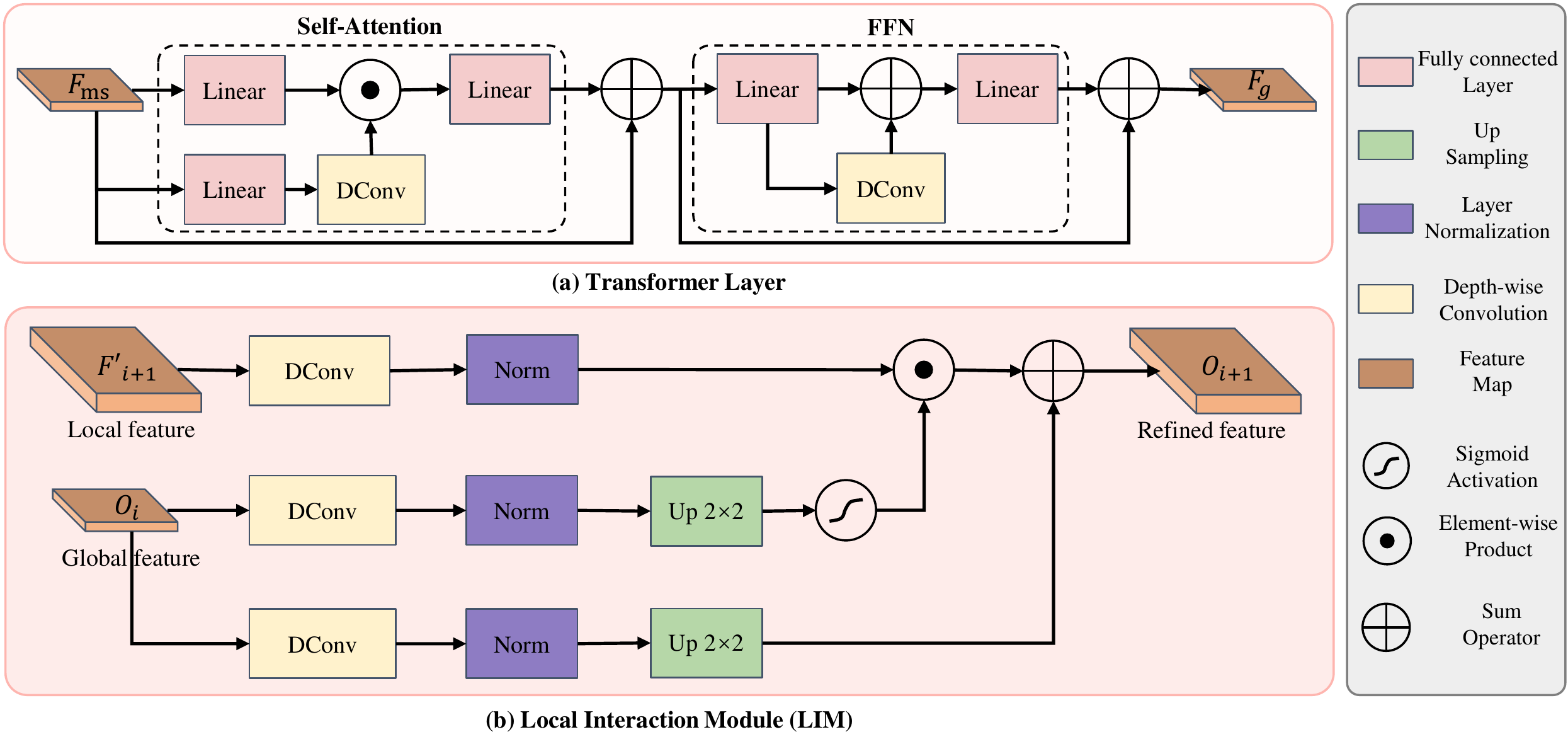}
\caption{Structure diagram of the transformer layer and local interaction module. (a) Transformer Layer: we adopt a simple self-attention implemented by using a large kernel convolution (the kernel size of ``DConv" in self-attention is set to 11$\times$11) to obtain long-range contextual information; the self-attention is immediately followed by a feed-forward network (FFN). Some details are omitted for simplicity, \textit{e.g.}, reshaping the feature maps. (b) Local Interaction Module (LIM): we use convolutional modulation to fuse local features from skip connections and global features from the transformer layer to obtain a refined feature.}
\label{fig:transformer+lim}
\end{figure*}

The global feature $\mathbf{O}_{i}$ has a larger receptive field and contains high-level information. In contrast, the local feature $\mathbf{F}'_{i+1}$ contains low-level texture information, but the receptive field is limited. We design a local interaction module (LIM) to fuse these two features effectively. The local interaction module (see Figure~\ref{fig:transformer+lim}(b)) is the core component in the decoder (see Figure~\ref{pmaa}(c)), which gradually restores image resolution through the previously modulated features $\mathbf{F}_{i}'$

Specifically, we use $\mathbf{O}_{i}$ after upsampling as the weights of $\mathbf{F}'_{i+1}$ to obtain more robust local features. At the same time, $\mathbf{O}_{i}$ are convolutionally modulated and then residual concatenated with $\mathbf{F}'_{i+1}$ to obtain a refined feature representation $\mathbf{O}_{i+1}$ containing both global and local information:
\begin{equation}
    \mathbf{O}_{i+1 }=\phi(\sigma(\mathbf{D}_{1}(\mathbf{O}_{i}))) \odot \mathbf{D}_{2}(\mathbf{F}_{i+1}')+\phi(\mathbf{D}_{3}(\mathbf{O}_{i})),
\end{equation}
where $\mathbf{D}_{1},\mathbf{D}_{2},\mathbf{D}_{3}$ are three depth-wise convolutional layers, $\sigma$ represents the sigmoid activation function, and $\phi$ represents the nearest-neighbor interpolation, some details such as the normalization layer are omitted. Finally, we shall take the feature map with the highest resolution as the output of the current cloud removal model.

\subsection{Progressive learning}

Progressive learning (PL) has been introduced to image inpainting~\cite{repaint,eii} and image restoration~\cite{mprnet,restormer} tasks and has achieved superior performance. Multi-temporal cloud removal is a complex task between image inpainting and restoration. Instead of attempting a direct transformation from multi-temporal cloudy images to a single cloud-free image, we adopt a strategy of breaking down the cloud removal process into several smaller, more tractable steps. These preliminary stages assist in laying the foundation for the latter stages of the training process:
\begin{equation}
    \mathbf{Q}_{t+1}=S(\mathbf{Q}_{t} \oplus \varphi(\mathbf{U_{c}})),
\end{equation}
where $\mathbf{Q}_{t}$ denotes the output of the cloud removal module at time $t$,  $S$ represents a cloud removal module, and $\varphi$ represents a convolutional layer with kernel size of $1 \times 1$ to do the work of channel transformation. Although too many stages may cause the network to be too deep to converge, increasing the model's parameters and computational cost. We explored the effects of different stages in the ablation experiments (see Figure~\ref{fig:pls}). We chose the optimal number of stages to balance performance and efficiency.

\subsection{Loss function} 

The cloud removal network is optimized to improve the model's ability to eliminate clouds and achieve high-fidelity cloud-free image generation. 
To align the divergence between the generated cloud-free image and the ground truth, we calculate the L1 loss between them
\begin{equation}
    \mathcal{L}_\text{L1}(F)=||\mathbf{y}-F(x)||_{1},
\end{equation}
where $x$ represents multi-temporal cloudy images, $F$ represents cloud removal model, $y$ represents the ground truth, and $F(x)$ denotes the estimated cloud-free image.

\section{Experiments}

\subsection{Datasets}

To validate the efficacy of PMAA and its novel components, we conduct experiments on two widely-recognized cloud removal datasets.

\textbf{\emph{Sen2\_MTC\_Old}}. This dataset~\cite{stgan} is created from publicly-available Sentinel-2 images. It contains 945 different tiles with a total of 3130 image pairs. Every three cloudy images correspond to one cloud-free image, where each image has size $(w, h) = (256, 256)$, the number of channels $C = 4$ (RGB and infra-red), and pixel values in the range $[0, 255]$. The pixel values of each image are normalized to $[0, 1]$ before input and then transformed to $[-1, 1]$ by a mean and variance of 0.5. This dataset divides into a training set, a validation set, and a test set in the ratio of $8:1:1$.

\textbf{\emph{Sen2\_MTC\_New}}. This dataset~\cite{ctgan} is also created from the Sentinel-2 images, which contains about 50 non-overlapping tiles with about 70 pairs of images per tile. Its settings are consistent with the \emph{Sen2\_MTC\_Old} dataset, except that the pixel value range is $[0, 10000]$. Compared to the \emph{Sen2\_MTC\_Old} dataset, it has higher resolution and annotation quality. This dataset divides into training, validation, and test sets in the ratio of $7:1:2$.

\subsection{Implementation details}

\textbf{Training settings}. Initially, we normalize all images to [-1, 1], then concatenate multiple cloudy images along the channel dimension, feeding them into several bottleneck layers consisting of convolutions to extract features. The downsampling and upsampling counts of the encoder and decoder are set to 4. The channel count of the hidden layer is set to 32. Finally, PMAA predicts cloud-free image through a convolution layer with a kernel size of 3$\times$3 and a stride of 1$\times$1. During the training phase, we use AdamW~\cite{adamw} optimizer with an initial learning rate of $5\times 10^{-4}$ and weight decay $1\times 10^{-5}$, and a cosine decay~\cite{cosine} learning rate schedule. We train all models for 100 epochs, with a batch size of 4, and save the model with the best SSIM value on the validation set for testing on the test set. We conduct experiments on a machine with 4 $\times$ NVIDIA GeForce RTX 3090 (24GB memory).

\textbf{Evaluation metrics}. In all experiments, we report the Peak Signal-to-Noise Ratio (PSNR, dB) and Structual Similarity Index Measure (SSIM~\cite{ssim}) of the test set to evaluate the precision of the generated cloud-free images. To evaluate model efficiency, we report the number of parameters (M) and multiply–accumulate operations (MACs, G) for all models. The result is calculated via the ptflops\footnote{https://github.com/sovrasov/flops-counter.pytorch}.

\begin{table*}[t]
\centering
\setlength{\tabcolsep}{10pt}
\caption{Ablation studies of MAM and LIM on the \emph{Sen2\_MTC\_New} dataset. ``$\checkmark$" indicates that this part is used, while ``$\times$" denotes the absence of its use.}
\label{tab:ablation}
\begin{tabular}{cccccccccc}
\toprule
\multicolumn{5}{c}{MAM}                                                                                               & \multirow{3}{*}{LIM} & \multirow{3}{*}{PSNR $\uparrow$} & \multirow{3}{*}{SSIM $\uparrow$} & \multirow{3}{*}{Params (M) $\downarrow$} & \multirow{3}{*}{MACs (G) $\downarrow$} \\ \cline{1-5}
\multicolumn{2}{c}{Multi-scale fusion} & \multicolumn{2}{c}{Transformer layer} & \multirow{2}{*}{Selective attention} &                      &                       &                       &                         &                       \\ \cmidrule(r){1-2} \cmidrule(r){3-4}
Concat              & Sum              & Patch            & Nonpatch           &                                      &                      &                       &                       &                         &                       \\ \midrule
$\times$                           & $\times$                          & $\times$                           & $\times$                           & $\times$                                       & $\times$                         & 17.441                   & 0.581                    & 2.71                           & 90.30                        \\
\cellcolor[HTML]{FFCCC9}$\times$   & \cellcolor[HTML]{FFCCC9}$\times$  & $\times$                           & $\checkmark$                           & $\checkmark$                                       & $\checkmark$                         & 18.201                   & 0.601                    & 3.44                           & 91.91                        \\
\cellcolor[HTML]{FFCCC9}$\checkmark$   & \cellcolor[HTML]{FFCCC9}$\times$  & $\times$                           & $\checkmark$                           & $\checkmark$                                       & $\checkmark$                         & 18.045                   & 0.606                    & 3.73                           & 92.01                        \\
$\times$                           & $\checkmark$                          & \cellcolor[HTML]{FFCE93}$\times$   & \cellcolor[HTML]{FFCE93}$\times$   & $\checkmark$                                       & $\checkmark$                         & 17.809                   & 0.589                    & 2.83                           & 91.78                        \\
$\times$                           & $\checkmark$                          & \cellcolor[HTML]{FFCE93}$\checkmark$   & \cellcolor[HTML]{FFCE93}$\times$   & $\checkmark$                                       & $\checkmark$                         & 18.032                   & 0.599                    & 3.25                           & 91.85                        \\
$\times$                           & $\checkmark$                          & $\times$                           & $\checkmark$                           & \cellcolor[HTML]{FFFC9E}$\times$               & $\checkmark$                         & 18.300                   & 0.608                    & 3.43                           & 91.87                        \\
$\times$                           & $\checkmark$                          & $\times$                           & $\checkmark$                           & $\checkmark$                                       & \cellcolor[HTML]{CBCEFB}$\times$ & 17.590                   & 0.597                    & 3.32                           & 90.56                        \\
\cellcolor[HTML]{FFCCC9}$\times$   & \cellcolor[HTML]{FFCCC9}$\checkmark$  & \cellcolor[HTML]{FFCE93}$\times$   & \cellcolor[HTML]{FFCE93}$\checkmark$   & \cellcolor[HTML]{FFFC9E}$\checkmark$               & \cellcolor[HTML]{CBCEFB}$\checkmark$ & \textbf{18.369}          & \textbf{0.614}           & 3.44                           & 91.94                        \\ \hline
\end{tabular}
\end{table*}

\subsection{Ablation studies}\label{sec:ablation}

To quantitatively assess the individual impact of each component, we conduct a series of extensive ablation experiments on the \emph{Sen2\_MTC\_New} dataset, which possesses superior annotation quality compared to the \emph{Sen2\_MTC\_Old} dataset.

\begin{figure}[t]
\centering
\includegraphics[width=\linewidth]{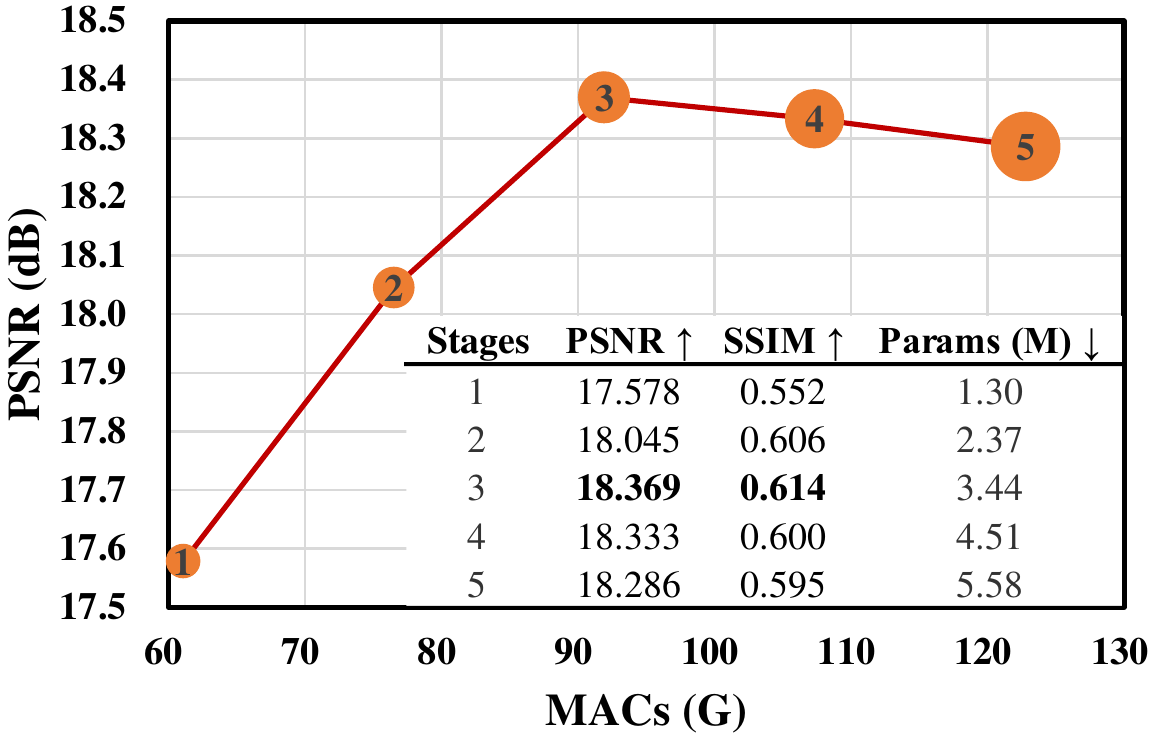}
\vspace{-5mm}
\caption{Quantitatively compare the impact of the number of progressive learning stages. The area of the circle indicates the number of model parameters, whereas larger circles indicate more parameters. The numbers in circles indicate the number of different stages.}
\label{fig:pls}
\end{figure}

\begin{figure*}[ht]
\centering
\includegraphics[width=\linewidth]{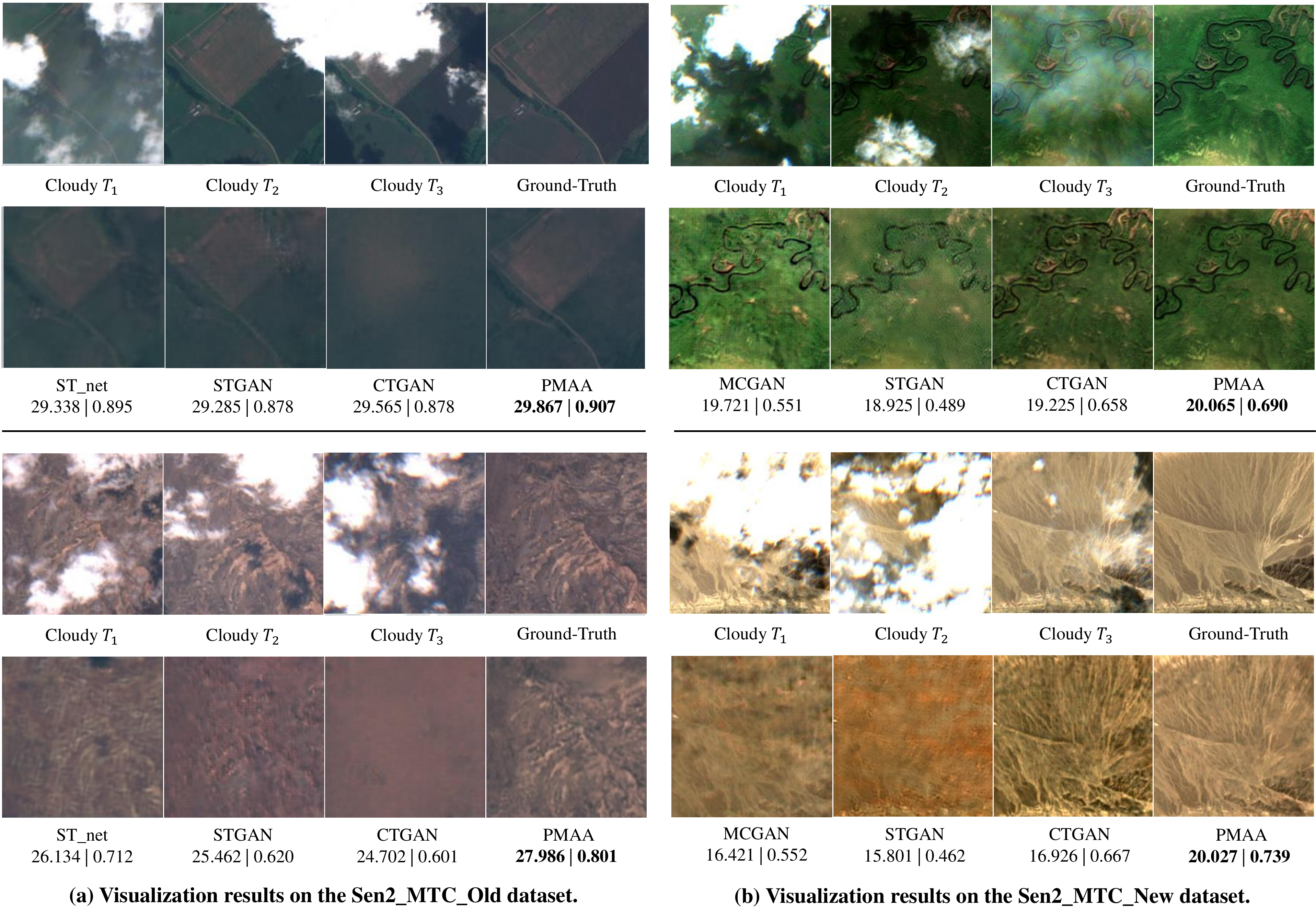}
\caption{Visualization results on \emph{Sen2\_MTC\_Old}  and \emph{Sen2\_MTC\_New}. The left and right sides of ``$|$" indicate the values of PSNR and SSIM, respectively.}
\label{fig:visualization}
\end{figure*}

\begin{table*}[t]
\centering
\setlength{\tabcolsep}{19.5pt}
\caption{Quantitative comparison of cloud removal performance and efficiency between PMAA and existing models on \emph{Sen2\_MTC\_Old} and \emph{Sen2\_MTC\_New}.}
\label{tab:pcr}
\begin{tabular}{ccccccc}
\toprule
\multirow{2}{*}{Methods} & \multicolumn{2}{c}{\emph{Sen2\_MTC\_Old}} & \multicolumn{2}{c}{\emph{Sen2\_MTC\_New}} & \multirow{2}{*}{Params (M) $\downarrow$} & \multirow{2}{*}{MACs (G) $\downarrow$} \\
\cmidrule(r){2-3} \cmidrule(r){4-5} 
                         & PSNR $\uparrow$                   & SSIM $\uparrow$                 & PSNR $\uparrow$              & SSIM $\uparrow$            &                         &                       \\
                         \midrule
MCGAN~\cite{mcgan}                    & 21.146                 & 0.481                & 17.448             & 0.513            & 54.42                   & 71.56                 \\
Pix2Pix~\cite{pix2pix}                  & 22.894                 & 0.437                & 16.985             & 0.455            & 11.41                   & 58.94                 \\
AE~\cite{ae}                       & 23.957                 & 0.800                & 15.251             & 0.412            & 6.53                    & \textbf{35.72}                 \\
ST\_net~\cite{stnet}                  & 26.321                 & 0.834                & 16.206             & 0.427            & 4.64                    & 304.31                \\
STGAN~\cite{stgan}             & 26.186                 & 0.734                & 18.152             & 0.587            & 231.93                  & 1094.94               \\
CTGAN~\cite{ctgan}                    & 26.264                 & 0.808                & 18.308             & 0.609            & 642.92                  & 632.05                \\
\midrule
\rowcolor[RGB]{217,217,217} PMAA (Ours)                     & \textbf{27.377}                 & \textbf{0.861}                & \textbf{18.369}             & \textbf{0.614}            & \textbf{3.44}                    & 91.94                 \\
\bottomrule
\end{tabular}
\end{table*}

\subsubsection{Multi-scale fusion strategy}

We investigate the consequences of various multi-scale feature fusion approaches on cloud removal performance, as depicted in Table~\ref{tab:ablation}. In contrast to abstaining from feature fusion (utilizing the feature map with the most reduced resolution as the transformer layer input), the summation operation emerges as the optimal feature fusion strategy, culminating in a PSNR enhancement of 0.168 and SSIM enhancement of 0.013. It is worth noting that unsuitable feature fusion strategies (such as channel-dimension concatenation) may yield not only inferior results but also elevate computational complexity.

\subsubsection{Self-attention implementation}

As outlined in Table~\ref{tab:ablation}, we examine the impact of various self-attention strategies on cloud removal performance. Compared to the transformer layer based on patch (like Swin-Transformer~\cite{swin}), the nonpatch-based transformer layer (PMAA's transformer layer) obtains better performance, with PSNR and SSIM rising by 0.337 and 0.015, respectively. One possible explanation for this substantial advancement could be that the patch operation may lead to some patches lacking cloud occlusion while others being entirely cloud occlusion, thus rendering PMAA incapable of learning efficiently. 


\subsubsection{Selective attention} 

We also investigate the impact of the selective attention module in MAM on cloud removal performance, as shown in Table~\ref{tab:ablation}. Note that when the selective attention module is not included in the model, the skip connections are consistent with those of U-Net \cite{unet}. The results demonstrate that utilizing the selective attention module leads to improved cloud removal performance (0.069 increase in PSNR and 0.006 increase in SSIM), while introducing an extremely low parameter and computational complexity (approximately 0.01M and 0.07 GMACs). This can be explained by the fact that shallow feature extraction in the network contains some redundant information, whereas the adoption of selective attention mechanisms can effectively filter out such useless information while enhancing the useful information, resulting in a more refined feature representation.

\subsubsection{Local interaction module} 

Furthermore, we examine the ramifications of incorporating the LIM within the decoder on cloud removal efficacy. As delineated in Table~\ref{tab:ablation}, leveraging the Local Interaction Module yields superior performance (0.779 and 0.017 increase in PSNR and SSIM, respectively) in comparison to its absence (as observed in the initial upsampling operation in U-Net~\cite{unet}). This underscores the significance of local feature amalgamation in directing attention toward pertinent areas.

\subsubsection{Progressive learning} 

Expanding the quantity of cloud removal autoencoders augments the progressive learning procedure, consequently boosting the model’s expressive capabilities. Nonetheless, this simultaneously incurs additional computational expenses. In Figure~\ref{fig:pls}, we scrutinize the influence of the stage count in PMAA on model efficacy and performance. A satisfactory equilibrium between performance and efficiency is attained when the stage count within PMAA is assigned a value of 3. If the stage count perpetually increases, the model exhibits excessive intricacy and fails to converge, while the cross-stage transfer may precipitate information attrition, culminating in suboptimal results. Consequently, we adopt a default configuration comprising three stages in the conducted experiments.

\subsection{Comparison with the state-of-the-arts}

We conduct extensive experiments to compare the performance and efficiency of our proposed PMAA with existing models on \emph{Sen2\_MTC\_Old} and \emph{Sen2\_MTC\_New} datasets. Our model achieves SOTA cloud removal performance using a very small number of parameters, which can be called a lightweight network.

\textbf{Performance of cloud removal.} we present a comprehensive comparison of the quantitative results of our proposed method with existing state-of-the-art methods on the \emph{Sen2\_MTC\_Old} and \emph{Sen2\_MTC\_New} datasets, as detailed in Table~\ref{tab:pcr}. On both datasets, PMAA achieves consistent SOTA performance on PSNR and SSIM, demonstrating our method's superiority. On two benchmark datasets, PMAA performs much better than the previous SOTA model CTGAN and other methods with only 0.5\% of the number of parameters compared to CTGAN. 

The visualization results are shown in Figure~\ref{fig:visualization}, which presents a qualitative comparison of our method against existing approaches on representative examples from two datasets. We observe that PMAA can consistently generate more detailed structures and demonstrates improved robustness in cloud removal on two datasets.

\textbf{Efficiency of cloud removal.} We compare with existing methods regarding the number of model parameters and MACs (reflecting the computational complexity). In Table~\ref{tab:pcr}, we observe that PMAA has a superior efficiency with 0.5\% and 14.6\% of the previous SOTA model CTGAN's number of parameters and MACs, respectively. Our efficiency gain is attributed to two components: MAM and LIM. First, the encoder uses depth-wise convolution layers for downsampling, and MAM fuses multi-scale features using parameter-free pooling layers. This is more efficient than CTGAN, which distributes the computational cost in each layer. Second, our transformer layer in MAM does not compute the similarity matrix and uses only the large convolutional kernel to extract features since it calculates at the minimum resolution scale, which is sufficient to extract global features. Third, LIM in the decoder uses very few parameters to fuse global and local features to obtain refined feature representations. The extremely low parameter number and computational complexity suggest PMAA's excellent deployability to low-resource devices.

\section{Conclusion}

To address the issue of low efficiency in current cloud removal methods, this paper presents a high-performance network refer as PMAA for cloud removal. PMAA leverages the autoencoder architecture with an efficient multi-scale attention module to capture global contextual information and guide the local interaction module in the decoder to reconstruct cloud-free images. Meanwhile, PMAA uses progressive learning to improve the model performance further. Sufficient experiments demonstrate that PMAA has essential implications for grounding the cloud removal model in practical deployments and is expected to achieve superior performance in large-scale and batch-processing tasks. In addition, owing to the efficacy and lightweight, we believe that the modules we have introduced can be seamlessly transplanted to other related problems, such as image restoration (e.g., rain, snow, fog, and noise removal) and other generative tasks.

\ack This work was supported in part by the Natural Science Foundation of China under Grant No. 62222606 and 62076238, in part by the Research on Efficiency Design of 3D Virtual Interactive Scene (k992146), and in part by the Research Foundation of the Key Laboratory of Spaceborne Information Intelligent Interpretation.

\bibliography{ecai}
\end{document}